 \documentclass[pmlr,twocolumn,10pt]{jmlr} 

\usepackage{booktabs}
\usepackage{siunitx}
\usepackage{graphicx}
\usepackage{float}
\usepackage{graphicx}
\usepackage{afterpage} 
\usepackage{multirow}
\usepackage{tikz}
\usepackage{adjustbox}
\usepackage{hyperref}
\usepackage{wrapfig}
\usepackage{graphicx}
\usepackage{url}
\usepackage{mdframed}

\newcommand{\equal}[1]{{\hypersetup{linkcolor=black}\thanks{#1}}}

\theorembodyfont{\upshape}
\theoremheaderfont{\scshape}
\theorempostheader{:}
\theoremsep{\newline}

\jmlrvolume{LEAVE UNSET}
\jmlryear{2023}
\jmlrsubmitted{LEAVE UNSET}
\jmlrpublished{LEAVE UNSET}
\jmlrworkshop{Machine Learning for Health (ML4H) 2023} 

 \title[MemoryCompanion]{MemoryCompanion: A Smart Healthcare Solution to Empower Efficient Alzheimer's Care Via Unleashing Generative AI}

\author{%
\Name{Lifei Zheng}\equal{These authors contributed equally} \Email{lz2278@nyu.edu}\\
\addr New York University Abu Dhabi, UAE
\AND
\Name{Yeonie Heo}\footnotemark[1] \Email{sh5874@nyu.edu}\\
\addr New York University Abu Dhabi, UAE
\AND
\Name{Yi Fang} \Email{yfang@nyu.edu}\\
\addr New York University Abu Dhabi, UAE
}

\begin{document}
\maketitle
\begin{abstract}
With the rise of Large Language Models (LLMs), notably characterized by GPT frameworks, there emerges a catalyst for novel healthcare applications. Earlier iterations of chatbot caregivers, though existent, have yet to achieve a dimension of human-like authenticity. This paper unveils `MemoryCompanion' a pioneering digital health solution explicitly tailored for Alzheimer's disease (AD) patients and their caregivers. Drawing upon the nuances of GPT technology and prompt engineering, MemoryCompanion manifests a personalized caregiving paradigm, fostering interactions via voice-cloning and talking-face mechanisms that resonate with the familiarity of known companions. Using advanced prompt-engineering, the system intricately adapts to each patient's distinct profile, curating its content and communication style accordingly. This approach strives to counteract prevalent issues of social isolation and loneliness frequently observed in AD demographics. Our methodology, grounded in its innovative design, addresses both the caregiving and technological challenges intrinsic to this domain.
\end{abstract}

\begin{figure*}[!ht]
  \centering
  \includegraphics[width=\textwidth]{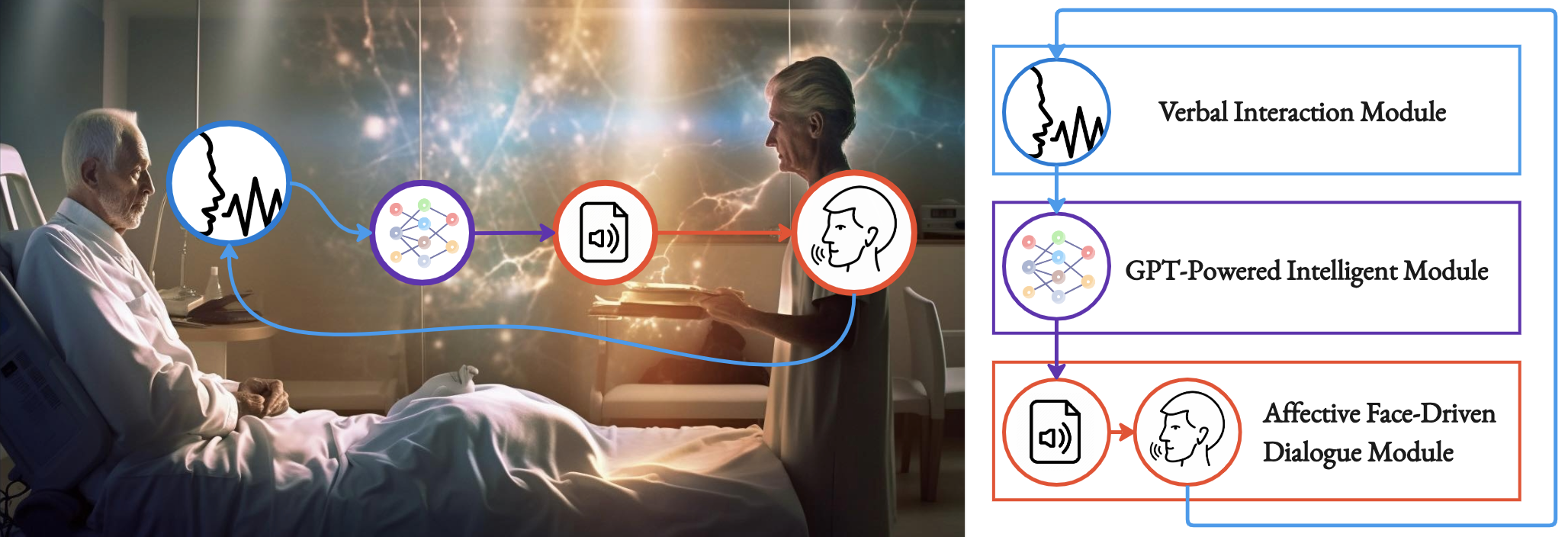}
  \caption{MemoryCompanion's Overview. The Verbal Interaction Module enables real-time vocal communication between AD patients and a GPT model, improving efficiency for those with cognitive deficits. The transcribed input of the patient is then processed into the tailored AD-Patient-GPT Language Model that supports in personalized conversations and reminiscence therapy, enhancing cognitive function and daily communication for AD patients in GPT-Powered Intelligent Module. Lastly, Affective Face-Driven Dialogue Module uses emotion-sensitive talking-face interface for authentic human-machine interaction, reducing feelings of loneliness and social isolation. Figure 1 is generated by~\cite{midjourney}.}
  \label{fig:overview}  
\end{figure*}

\section{Introduction}
Alzheimer's disease (AD) is the foremost neurodegenerative disorder~\cite{alzheimers-association-2023, niehs-2022}, impacting 1 in 9 individuals (10.8\%) aged 65 or older~\cite{ozela-2018}, with global prevalence projected to increase by 62\% by 2030, reaching approximately 47 million patients~\cite{rajan-2021}. Amidst this growing concern, present treatment options are restricted~\cite{ralph-2018}. Given the increased risk of stroke and death associated with antipsychotics in AD patients~\cite{maust-2015}, non-pharmacological approaches are prioritized to emphasize activities and social interactions for enhancing cognitive function and decelerating memory loss progression~\cite{who-2023, non-drug-2013}. However, the provision of continuous interaction and companionship for AD patients encounters serious hurdles of limited caregiver availability~\cite{kieboom-2020}, isolation, and daily socialization challenges~\cite{caregiver-stress-2023}. 

To alleviate the demanding physical and mental strains faced by caregivers, we introduce MemoryCompanion, a digital healthcare solution leveraging Generative Pre-trained Transformer (GPT) technology~\cite{OpenAI_API}. GPT serves as the core technology behind ChatGPT, enabling MemoryCompanion to be a continuous companion for AD patients. MemoryCompanion integrates GPT with multimedia technologies, including speech-to-text transcription~\cite{Hannun-2014}, voice cloning~\cite{coqui-tts}, and talking-face~\cite{StyleTalk}, to establish an efficient caregiving system for AD patients. The integration of voice cloning and talking-face technologies within GPT responses offers AD patients the sensation of communicating with familiar individuals at their convenience. As a robust caregiving alternative, MemoryCompanion consistently delivers care, mitigating the challenges encountered by human caregivers~\cite{Ohno2021, Koskie2018}. MemoryCompanion's scope extends beyond AD, yielding benefits across diverse health and educational domains, thus generating a broader social impact. 

Figure ~\ref{fig:overview} presents the architectural blueprint of MemoryCompanion, focusing on three integral modules: the Verbal Interaction Module, the GPT-Powered Intelligent Module, and the Affective Face-Driven Dialogue Module. Together, these components facilitate a unified and user-friendly experience for AD patients. In this study, our \underline{contributions} are delineated in three pivotal aspects, each of which highlights the significant advancement brought forth by MemoryCompanion:

\noindent\textbf{Patient-Centric Language Learning Model}
We introduce an innovative patient-centric Language Learning Model (LLM) designed for engaging AD patients in not only day-to-day conversations and offering emotional support but assisting in memory retrieval based on their personal experiences. This approach contributes to personalized healthcare technologies, specifically engineered to tailor the distinctive requirements of AD patients, thereby significantly enhancing the quality of care provided.

\noindent\textbf{Structured AD Patient Profile Paradigm}
To tackle the issues of limited data availability and data sensitivity in smart health solutions for AD patients, we create a well-structured AD patient profile paradigm. Simultaneously, we construct a synthetic dataset, serving as a foundational resource for future research endeavors. This initiative fulfills the imperative for extensive data in this domain while fostering future developments in AD patient support technologies and research.

\noindent\textbf{AD Caregiving Evaluation Metrics}
We present a tailored set of metrics for evaluating the effectiveness of smart caregiving systems for AD patients. These metrics establish a rigorous benchmark for future research, facilitating systematic assessments of system reliability and effectiveness. This contribution establishes a standardized framework for advancing smart caregiving technologies, thereby enhancing the quality of care and support for AD patients and their caregivers.

\begin{figure*}[!ht]
  \centering
  \vspace*{-1cm}
\includegraphics[width=\textwidth]{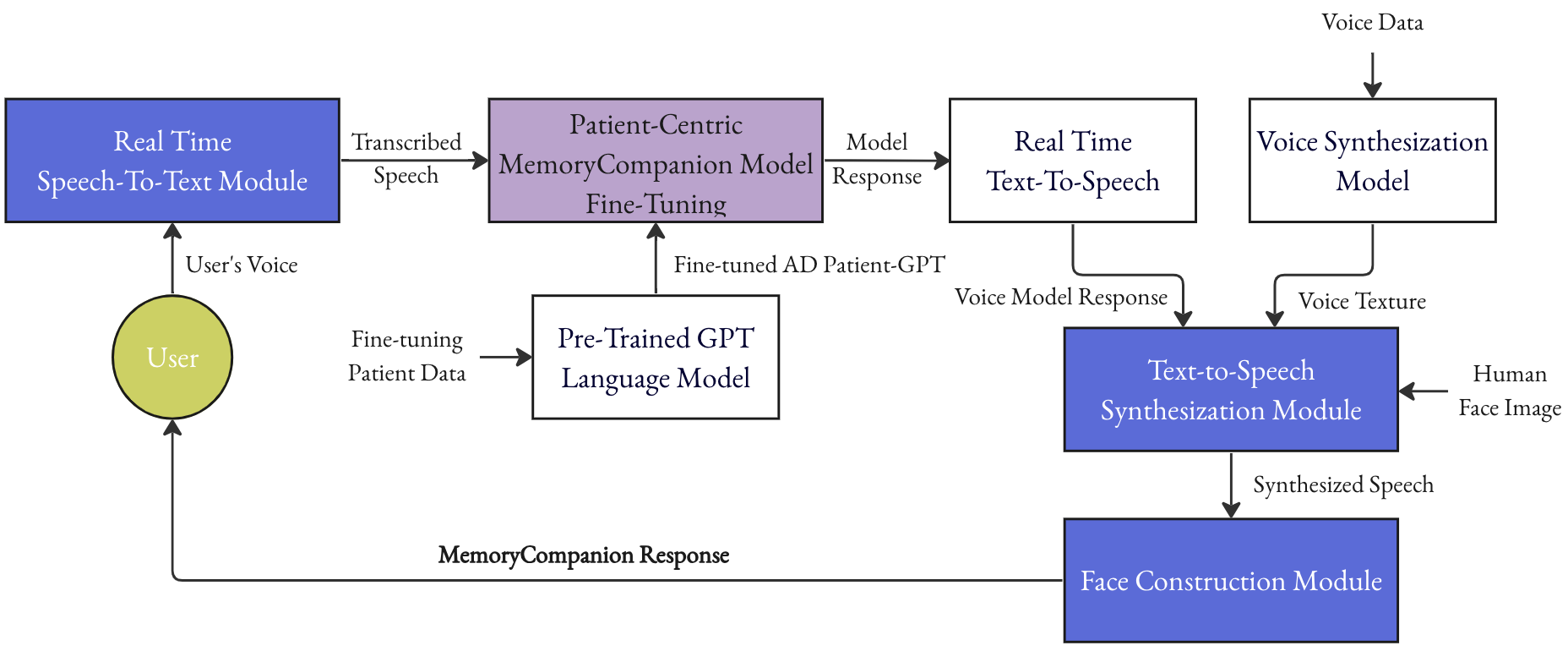}
  \caption{Methodology Overview. The four color-highlighted boxes represent the principal modules within our pipeline. Detailed descriptions of each module can be found in Modules \ref{section1} Real-Time Speech-to-text Recognition, \ref{section2} Patient-Centric MemoryCompanion
Model Fine-Tuning, \ref{section3} Text-to-Speech Synthesization, and \ref{section4} Talking Face Construction.}
  \label{fig2}
\end{figure*}

\vspace{-1em}
\section{Related Work}
\vspace{-1em}
Many smart health solutions have been developed to address the complex caregiving needs of patients with AD and on a broader scale, dementia. Prior studies were mostly developing applications that engaged in the domain of patient monitoring, health care service support, and patient education. Since 2012, Robotics-based applications have been investigated for assisting AD patients in tasks such as food preparation and consumption~\cite{miguelvaz-2020}, and participating in recreational activities within care homes~\cite{Yue-2015}. These robotic innovations have demonstrated considerable potential in mitigating psychological distress among AD patients and their caregivers by reducing agitation, depression, anxiety, impulsiveness, restlessness, and chronic pain through ``robotherapy"~\cite{Bhargava-2022}. However, there was a noticeable paucity of applications concerning caregiving skills and activities due to constraints in artificial intelligence and achieving an authentic human-like interaction~\cite{vicuna-2023}. 
\setlength{\parskip}{0pt}
\setlength{\parindent}{1.5em}

The introduction of Large Language Models (LLMs) has catalyzed research to overcome these limitations and enhance the human-like interactivity of these systems. Studies exploring the use of LLM-based chatbots, taking into account user personality profiles and emotional states to support dementia caregivers, have become increasingly prevalent~\cite{Nasiri-2023}. Similar studies, for example, MultiMedQA, presented in Nature in 2023, propose a large language model encoded with clinical knowledge trained through prompt engineering strategies and clinically relevant datasets~\cite{Singhal-2023}. 

Yet, a gap in the literature remains with no reported solutions integrating both LLM and visual or audio components to create a fully immersive caregiving environment for AD patients. In response to this gap, we propose MemoryCompanion, a revolutionary autonomous technology that combines GPT-based models with speech-to-text transcription, voice-cloning, and talking-face technologies. This innovative approach represents a significant leap forward in the domain of AD patient care, offering a more genuine, human-like interaction that not only improves the quality of life but also mitigates the sense of social isolation.

\vspace*{-0.3cm}
\section{Methodology}
The principal purpose of MemoryCompanion is to develop a digital GPT agent that accurately replicates the user's beloved one's visage, tonal nuances, and linguistic style in order to prompt a natural conversation flow and maintain high authenticity in human-computer interaction. To achieve this objective, MemoryCompanion comprises four distinct components. The overview of the methodology, as illustrated in Figure~\ref{fig2}, illustrates the overall architecture of MemoryCompanion, with sections \ref{section1}, \ref{section2}, \ref{section3}, and \ref{section4} indicating each module. Each module and its submodules is explained in detail below.

\vspace*{-0.38cm}
\subsection{Real-Time Speech-to-text Recognition}\label{section1}
To enable direct user-chatbot interaction, Module \ref{section1} Real-Time Speech-to-text Recognition is developed to transcribe the speech from AD patients into text. This module processes an audio stream as input and transforms it into a sequence of words and punctuations. The transcribed text is then forwarded to Module \ref{section2}, a fine-tuned GPT Language model, which allows for smooth text-response processing. Our real-time speech recognition module is implemented using an approach similar to that described in the DeepSpeech paper~\cite{Hannun-2014}.

\vspace*{-0.38cm}
\noindent\subsection{Patient-Centric MemoryCompanion Model Fine-Tuning} \label{section2}
\begin{figure}[h]
  \vspace*{-0.5cm}
  \centering
  \includegraphics[width=1\linewidth, height=0.8\textheight, keepaspectratio]{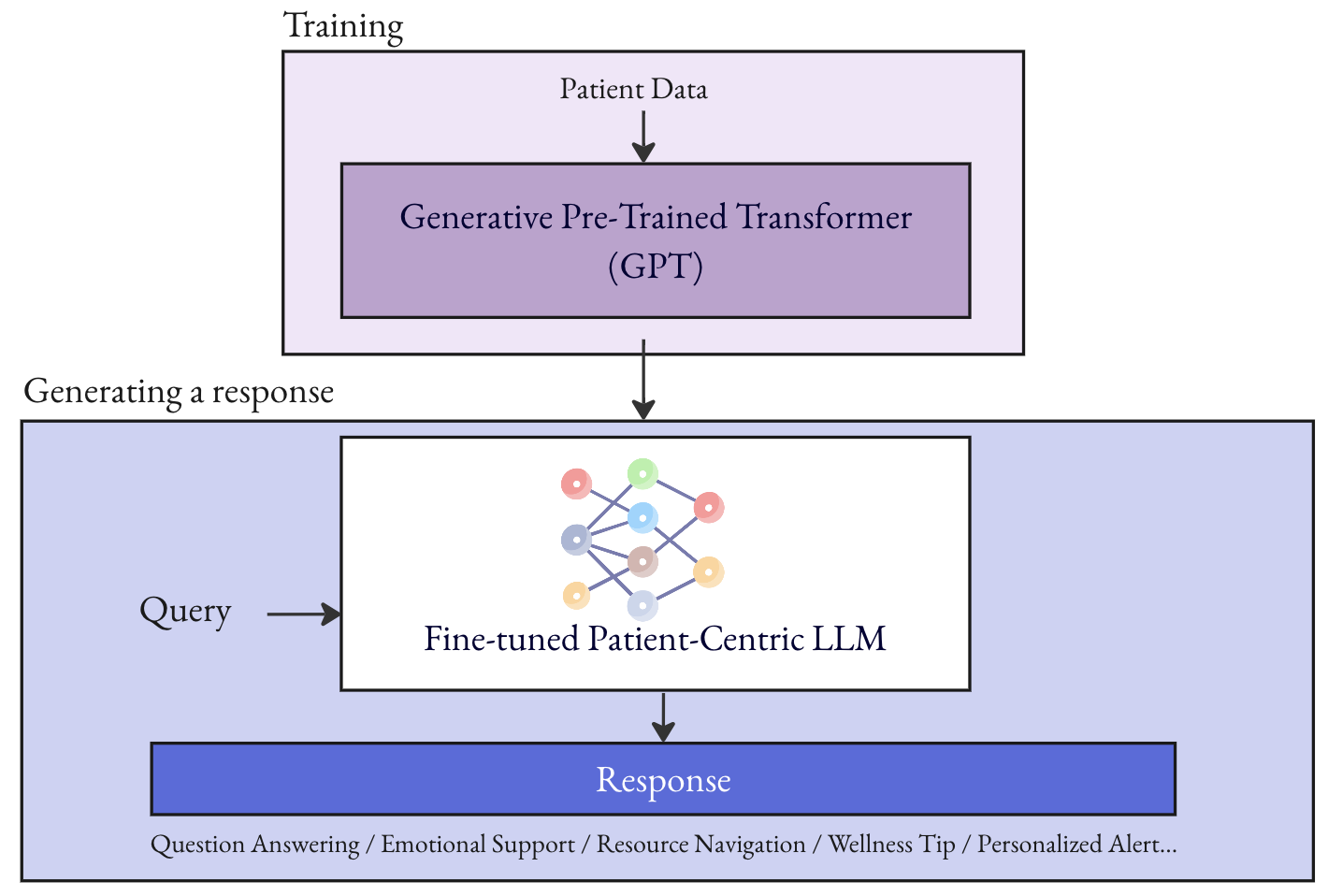}
  \caption{Patient-Centric LLM Fine-Tuning Diagram}
  \label{new_finetuning_diagram}
  \vspace*{-0.5cm}
\end{figure}

Upon receiving the transcribed speech from Module \ref{section1}, the subsequent stage entails processing the text and generating responses using our novel AD-patient-centric large language model. Our objective is to equip MemoryCompanion with ChatGPT-level \cite{OpenAI_API} language comprehension and processing capabilities, specifically fine-tuned to interact effectively with AD patients within a healthcare context.

To meet these goals of advanced language comprehensive and customized responses for AD patients, we develop a new MemoryCompanion Generative Language Model. This is achieved by fine-tuning the GPT-3.5 Turbo model~\cite{OpenAI_API} using domain-specific patient cases data. Through this fine-tuning process, we transition from a general-purpose GPT model to a highly specialized one, thereby enabling MemoryCompanion to comprehend the AD patients and generate text that is closely aligned with the unique requirements of these patients \cite{solulab2023}. As shown in Figure \ref{new_finetuning_diagram}, the fine-tuning process is described as:

\noindent\textbf{GPT Model Training}
The GPT-3.5 Turbo model is pretrained on a varied corpus. During this phase, the model's parameters, \( \theta \), are adjusted to optimize the log-likelihood of the observed data, as denoted by:

\vspace*{-0.2cm}
\begin{equation}
\theta^* = \arg \max_\theta \sum_{i=1}^N \log P(y_i | x_i, \theta)
\label{equation1}
\end{equation}

\begin{figure}[h]
\centering
\includegraphics[width=1\linewidth]{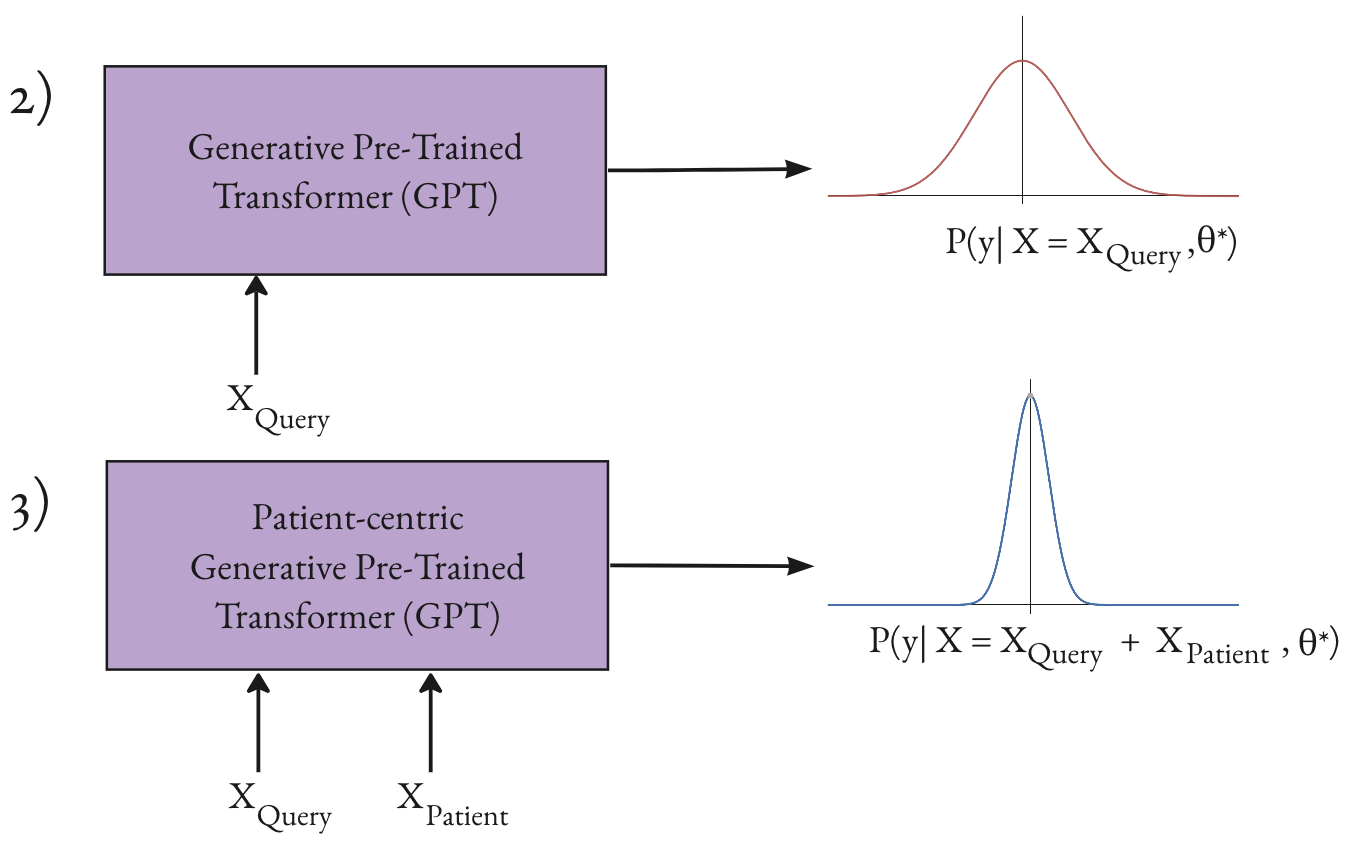}
\caption{GPT vs Patient-centric GPT}
\label{new_finetuning_diagram2}
\end{figure}

In this equation, \( x_i \) represents the input sequence, while \( y_i \) signifies the target output distribution. Given an input sequence, the model's response is generated by randomly sampling from this target distribution. This optimization seeks to find the configuration of \( \theta^*\) that makes the model’s predictions most consistent with the observed data.
Upon obtaining the optimal parameters \( \theta^* \), as described in Equation \ref{equation1} and part (2) in Figure \ref{new_finetuning_diagram2}, the model can generate responses by sampling from a probability distribution conditioned on a given query \( X_{\text{query}} \):

\begin{equation}
P(y | X = X_{\text{query}}, \theta^*)
\label{equation2}
\end{equation}

\noindent\textbf{Contextual Refinement for AD Patients} Despite the parameter optimization in the general GPT model in Equation \ref{equation2}, the resulting probability distribution may lack the requisite specificity for effective interactions with AD patients. To ensure tailored interactions, we implement prompt engineering, concatenating the query \( X_{\text{query}} \) with the patient profile \( X_{\text{patient}} \) as input sequence X, as visualized in Equation \ref{equation3} and part (3) in Figure \ref{new_finetuning_diagram2}:

\begin{equation}
P(y | X = X_{\text{query}} + X_{\text{patient}}, \theta^*)
\label{equation3}
\end{equation}

\begin{figure*}[h!]
  \vspace*{-0.9cm}
  \centering
\includegraphics[width=0.95\linewidth, height=0.95\textheight, keepaspectratio]
  {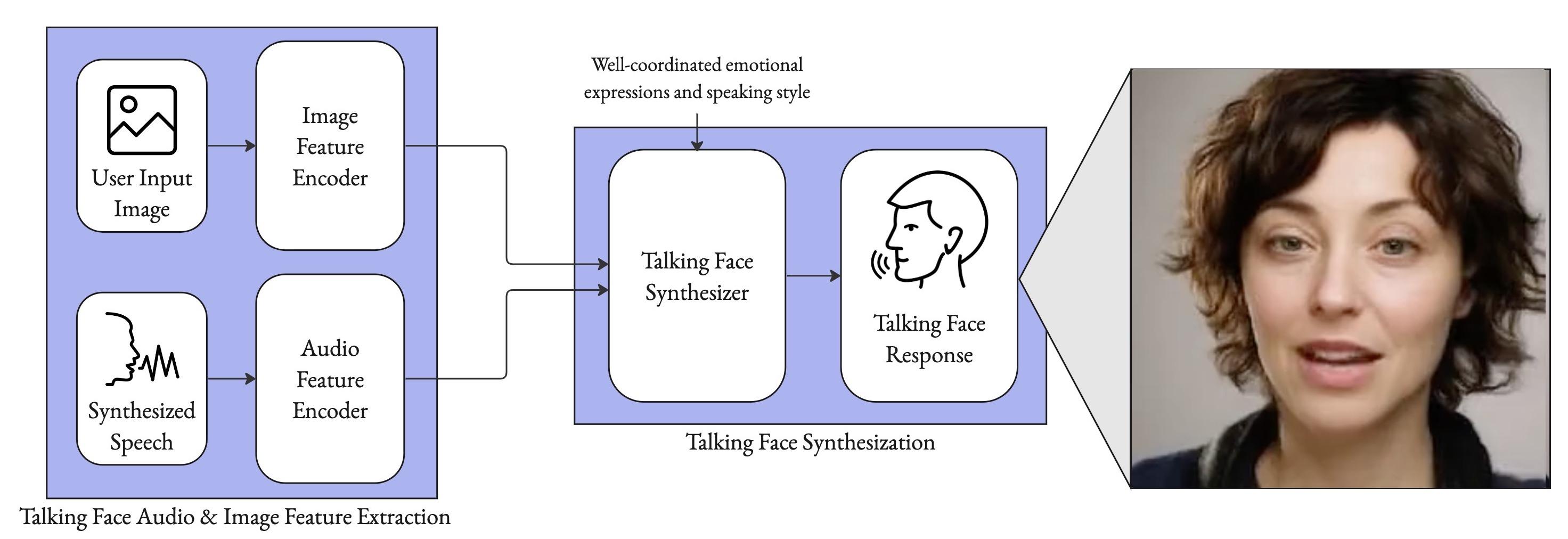}
  \caption{Talking Face Construction Diagram. The left side presents our talking face technology synthesization diagram. The right talking face figure presents a screenshot of our synthesized MemoryCompanion, designated as `Sam', delivering the output response from the patient-centric LLM model in an emotion-oriented and interactive way through our advanced talking-face multimedia technology. For a comprehensive visual representation of additional responses from MemoryCompanion, please refer to the supplementary videos in Appendix \ref{apd:first}. `Sam' is synthesized based on an image generated from~\cite{midjourney}.}
  \label{talking face}
\end{figure*}

This concatentation refines responses by leveraging the pretrained GPT model's knowledge and tailoring it to the specific conditions and queries of each AD patient. This process generates a more precise conditional probability distribution, reducing variance \( \sigma^2 \). This variance reduction enhances system efficacy, yielding focused, accurate, and contextually aware responses. The conditional probability distribution minimizes erratic or irrelevant outputs, critical in maintaining clarity and relevance with AD patients, while sharpening the model's precision, aligning output closely with the specific queries and conditions presented. This enhancement allows for accurate, contextually aware, and personalized responses that address the nuanced needs of AD patients. Notably, the efficacy of this approach is directly correlated with the diversity and granularity of the patient profile dataset. A more comprehensive dataset leads to a more refined and AD-patient-centric set of likely responses, underlying the significance of detailed patient profiles in this endeavor. 

\subsection{Text-to-Speech Synthesization} \label{section3}
To enhance user interaction, it's essential to convert text responses from the language model in Module \ref{section2} Patient-Centric MemoryCompanion Model Fine-Tuning into authentic speech. Module \ref{section3} Text-to-Speech Synthesization Module, refer to Figure \ref{fig2}, achieves this by utilizing Text-to-Speech (TTS) technology \cite{coqui-tts} which is designed expressly to transform text into speech. A standout feature is voice cloning, which mimics the conversational speech patterns of the beloved one chosen by the user, thus elevating the user experience. This voice cloning comprises two critical phases:

\noindent\textbf{Speaker Voice Reference Module}
This phase focuses on collecting targeted voice data to train a model that captures unique auditory characteristics of specific speakers, representing the characteristics as an embedding vector.

\noindent\textbf{Speech Synthesizer Module}
In this phase, we develop a synthesizer. Integrated with the embedding vector from last phase, it accurately translates textual input to speech. The paired text-audio datasets guarantee the synthesis of exceptional fidelity, which finally output a synthesized speech that will be used as input for Module \ref{section4} Talking Face Construction.

\subsection{Talking Face Construction} \label{section4}
Given the synthesized speech in Module \ref{section3}, the integration of a visually expressive talking face with emotional nuances becomes essential for enhancing user experience authenticity, as shown in Figure \ref{talking face}. Existing studies emphasize the importance of nonverbal cues and facial expressions in effective emotional communication \cite{Albert-1971, Michael-1971}. Observing a human figure with synchronized facial features bolsters communication effectiveness by reinforcing verbal discourse and tonality. Our methodology is inspired by StyleTalk~\cite{StyleTalk}. Our emotionally enriched talking face comprises two phases, as shown in Figure \ref{talking face}.

\noindent\textbf{Talking Face Audio \& Image Feature Extraction}
Synthesized speech, referring Figure \ref{talking face}, is transformed into word embeddings for vocal generation, while facial attributes from the user's input image, presented in Figure \ref{talking face}, are distilled into a vector. Both data forms merge through a self-attention pooling layer, resulting in the final style code.

\noindent\textbf{Talking Face Synthesization}
Utilizing style data from last phase, a standard transformer produces accurate lip syncing and authentic expressions. A talking face transformer, referring Figure \ref{talking face}, then adjusts the network, ensuring a talking face with well-coordinated emotional expressions and speaking style, enhancing overall user experience.

\begin{figure*}[!ht]
  \centering
  \adjustbox{frame=0.05pt 1pt}{\includegraphics[width=1\textwidth]{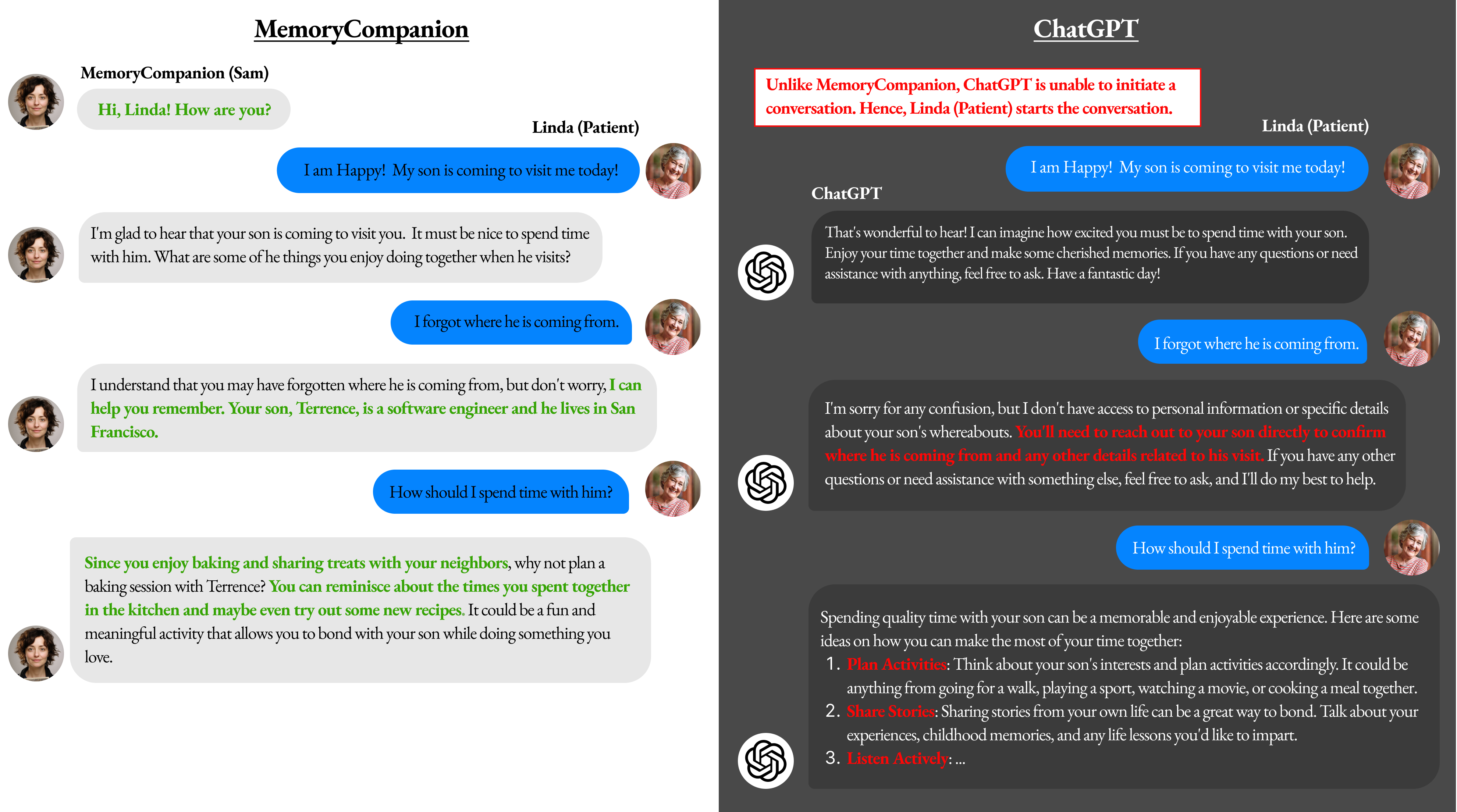}}
  \caption{Chat History of Linda with MemoryCompanion and ChatGPT}
  \label{chat4}
\end{figure*}


\section{Experiment \& Evaluation}
In the experiement section, we meticulously prepare the data our MemoryCompanion's ability to respond to patient queries, both qualitatively and quantitatively. After generating a synthetic patient case, Linda Williams in Figure \ref{chat4}, we qualitatively evaluate MemoryCompanion in comparison to OpenAI's ChatGPT 4.0, a general GPT model to highlight the unique contributions of our patient-centric approach. Next, various evaluation critera are employed to provide quantitative understanding of our system's performance for 100 synthetic patient cases. 

\subsection{Data Preparation}
The key to fine-tuning an AD patient language model lies in applying patient profile data. Recognizing the sensitivity and confidentiality of patient data, we have been cautious in our data collection. We sourced a dataset of profiles of AD patients from~\cite{BCMJ, NLM, Science_case_study}. Due to the limited availability of real patient data, we generate 10,000 synthetic patient cases whose structure follows a sourced dataset of profiles of AD patients from~\cite{BCMJ, NLM, Science_case_study}. The synthetic cases generation process consists of two phases.

\noindent\textbf{Patient Profile Case Structure Definition} The case structure starts with demographics like age and ethnicity, it transitions into medical history, familial background, and professional and personal pursuits. Notably, it delves into the patient's current challenges, future aspirations, and emotional well-being. This structured profile provides a holistic understanding of AD patients, amalgamating both quantitative and qualitative facets. It is imperative for ensuring nuanced and tailored interaction with AD patients. A detailed patient profile example can be found in Appendix \ref{apd:second}.

\noindent\textbf{Synthetic Cases Data Generation} After establishing the case structure framework, the subsequent step involves leveraging the ChatGPT-4 model \cite{OpenAIChat2023} to synthesize patient cases conforming to the specified structure. Utilizing the OpenAI ChatGPT platform, we fed this pre-defined case structure into the GPT-4 model, and instruct GPT model to comprehend the intricacies of the provided case structure and to generate 10,000 high-fidelity, available in Appendix \ref{apd:second}, realistic synthetic patient cases that are analogous to the original dataset~\cite{BCMJ, NLM, Science_case_study}.

\subsection{Qualitative Evaluation}
When evaluating interactions with AD patients, the tailored, patient-specific approach of MemoryCompanion in Figure \ref{chat4} presents distinct advantages over ChatGPT in three aspects in their responses. 

\noindent\textbf{Ability to Initialize Conversation}
Initiating dialogues fosters patient engagement and eases their cognitive load in sustaining interactions. MemoryCompanion proactively starts dialogues, evident from its greeting ``How are you?" to Linda in Figure \ref{chat4}. In contrast, ChatGPT tends to reactively await patient input, potentially making it harder for AD patients to maintain conversations.

\noindent\textbf{Provision of accurate and personalized information} 
Providing specific, accurate information is important for AD patients, ensuring clarity and reducing confusion.
MemoryCompanion effectively offers precise details about Linda's son, his residence and occupation, leveraging patient-specific data, as seen in Figure \ref{chat4}. Conversely, ChatGPT, without this granularity, generalizes its advice, potentially complicating matters for AD patients.

\noindent\textbf{Tailored responses for personalized interaction}
Beyond accuracy, personalized recommendations enhance the quality of care, addressing both emotional and cognitive needs of AD patients. In Figure \ref{chat4}, when Linda seeks suggestions about interacting with her son, MemoryCompanion integrates her interests, like baking, into its advice. This level of customization deepens the patient's connection and trust in the system's caregiving capabilities.

\subsection{Quantitative Evaluation}
Quantitative evaluation of MemoryCompanion is pivotal to validate its efficacy across diverse patient demographics. This ensures both broad applicability and tailored responsiveness in Alzheimer's care. Using criteria such as Accuracy, Ability To Lead Casual Conversation, Empathy \& Tone, and Error Handling, we examined its performance across varied, realistic patient interactions. From a dataset of 10,000 synthetic patient instances, we randomly selected 100 cases. Each was subjected to a question-answering process using nine tailored questions, according to Table \ref{questions}, designed to mirror real-world interactions MemoryCompanion might face, especially with AD patients. Responses were evaluated on a scale from 1 (ineffective) to 5 (exceptional).

\begin{table}[]
\begin{tabular}{|c|ll|}
\hline
\textbf{\begin{tabular}[c]{@{}c@{}}Evaluation\\ Criteria\end{tabular}}                                 & \multicolumn{2}{c|}{\textbf{Questions}}                                                                               \\ \hline
\multirow{3}{*}{Accuracy}                                                                              & \multicolumn{1}{l|}{Q1} & What is my name?                                                                            \\ \cline{2-3} 
                                                                                                       & \multicolumn{1}{l|}{Q2} & \begin{tabular}[c]{@{}l@{}}How is my family? Can \\ you talk a bit about them?\end{tabular} \\ \cline{2-3} 
                                                                                                       & \multicolumn{1}{l|}{Q3} & Where do I live?                                                                            \\ \hline
\multirow{3}{*}{\begin{tabular}[c]{@{}c@{}}Ability to \\ Lead \\ Casual \\ Conversations\end{tabular}} & \multicolumn{1}{l|}{Q4} & How are you today?                                                                          \\ \cline{2-3} 
                                                                                                       & \multicolumn{1}{l|}{Q5} & What should I do today?                                                                     \\ \cline{2-3} 
                                                                                                       & \multicolumn{1}{l|}{Q6} & \begin{tabular}[c]{@{}l@{}}How was the weather \\ today?\end{tabular}                       \\ \hline
\multirow{2}{*}{\begin{tabular}[c]{@{}c@{}}Empathy \& \\ Tone\end{tabular}}                            & \multicolumn{1}{l|}{Q7} & \begin{tabular}[c]{@{}l@{}}My friend passed away. \\ What should I do?\end{tabular}         \\ \cline{2-3} 
                                                                                                       & \multicolumn{1}{l|}{Q8} & \begin{tabular}[c]{@{}l@{}}I missed my family \\ members so much.\end{tabular}              \\ \hline
Error Handling                                                                                         & \multicolumn{1}{l|}{Q9} & \begin{tabular}[c]{@{}l@{}}Tell me about my time as \\ an astronaut?\end{tabular}           \\ \hline
\end{tabular}
\caption{Evaluation Criteria Questions}
\label{questions}
\end{table}

\begin{figure*}[ht]
  \centering
  \includegraphics[width=1\linewidth]{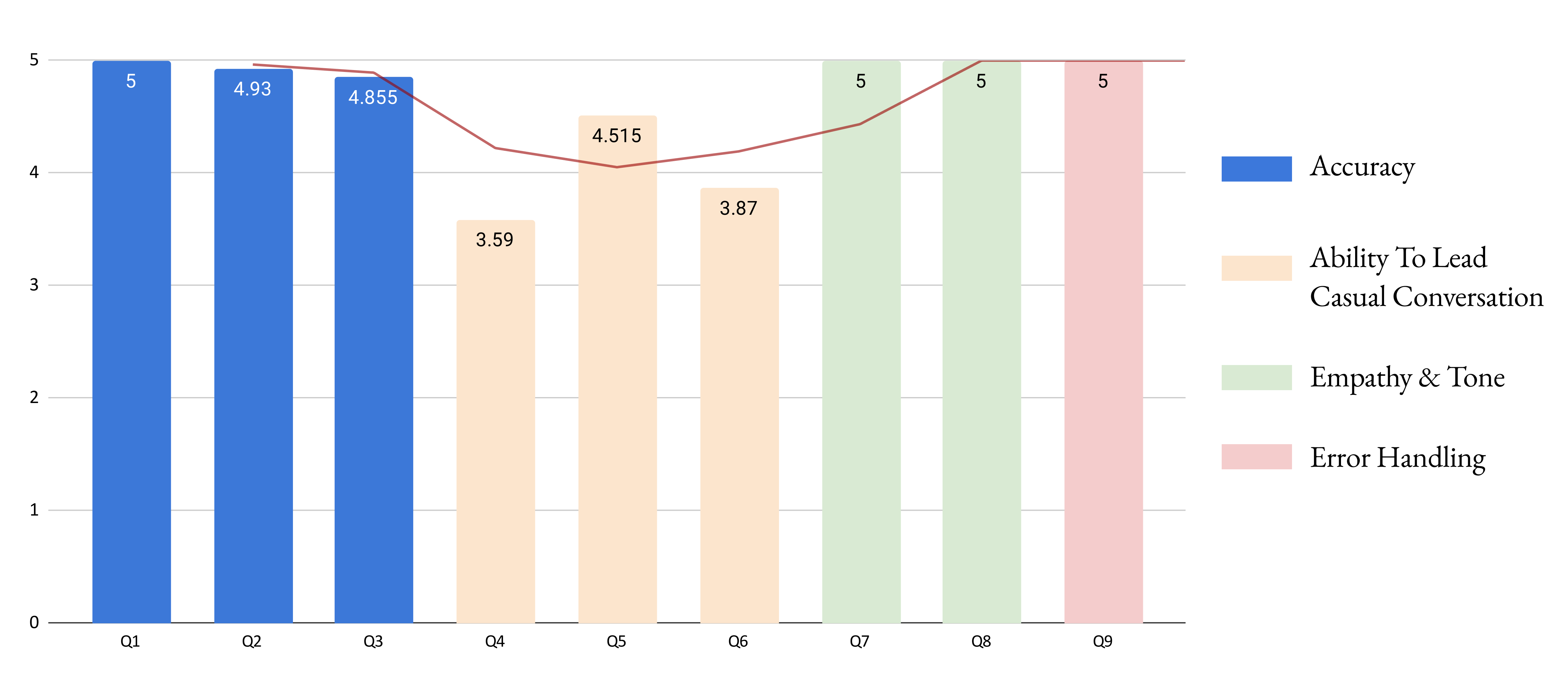}
  \caption{Average Ratings of Questions}
  \label{score_graph}
\end{figure*}

\vspace{-0.3em}
\subsubsection{Accuracy}


\vspace{-0.5em}
Accuracy is essential to MemoryCompanion, ensuring dependable information for AD patients. In Figure \ref{score_graph}, MemoryCompanion consistently delivers accurate details about the patient’s name, family, and residence. Variability in location-related responses remains acceptable, given the GPT model's complexities. However, while accurate, MemoryCompanion could enhance user experience by promoting more engaging continuity. For instance, asking about a patient's past, the origin of their name, or family activities might foster a more dynamic interaction, aligning with AD patients' and caregivers' needs.

\subsubsection{Ability to lead casual conversations}
\vspace{-0.4em}
This criterion evaluates MemoryCompanion's friendliness and its capability to create a natural and comforting interaction environment for AD patients. The ability to initiate and sustain casual conversations is of significant importance, as it contributes to the overall user experience and emotional well-being of the patients. As depicted in Figure \ref{score_graph}, MemoryCompanion offers warm greetings and recommends enjoyable activities, fostering a reassuring atmosphere. However, it occasionally falls short of achieving a completely human-like interaction, which is crucial for emulating conversations with familiar individuals. Importantly, disclosing its AI nature can disrupt the intended illusion. To ensure an optimal user experience, responses should consistently align with the expected personality of familiar individuals


\subsubsection{Empathy \& Tone}
\vspace{-0.4em}
Empathy and tone are essential criteria, particularly in the context of AD patients who may be dealing with sensitive and emotional challenges. MemoryCompanion's ability to demonstrate empathy and a supportive tone is crucial for providing emotional support and encouragement. This criterion ensures that the system can effectively acknowledge the feelings and address the needs of AD patients, enhancing their overall well-being and satisfaction. According to Figure \ref{score_graph}, MemoryCompanion excels in demonstrating empathy and a supportive tone in responses to sensitive and emotional questions. It consistently provides compassionate and comforting guidance to AD patients, acknowledging their feelings and offering valuable suggestions for coping with emotional challenges. 
\subsubsection{Error Handling}
\vspace{-0.4em}
Error handling is a pivotal criterion as it assesses MemoryCompanion's capacity to manage incorrect or out-of-scope questions. This capability is essential for maintaining a coherent and meaningful conversation with AD patients, who may occasionally make memory-related errors or ask questions beyond the system's scope. As seen from Figure \ref{score_graph}, MemoryCompanion demonstrates exceptional error handling in response to the out-of-scope question about being an astronaut. It accurately recognizes the error in the patient’s memory and provides gentle correction while offering meaningful details about the patient’s actual past as another professional occupation. 

\section{Limitation and Future Work}
\vspace{-0.2em}
While MemoryCompanion offers a groundbreaking approach to supporting AD patients, several crucial concerns arise:

\noindent\textbf{Ethical Dilemma} MemoryCompanion integrates familiar faces and tonal nuances of patients' loved ones, raising ethical questions about how patients perceive such intimate simulations. The emotional impact is profound, requiring a delicate balance between comfort and the risk of emotional dependency or memory distortion. Comprehensive ethical evaluations in collaboration with experts are essential to assess these design choices. A framework that prioritizes patients' emotional and cognitive well-being, ensuring the tool aids rather than distorts their genuine memories, is imperative.

\noindent\textbf{Data Equilibrium} The efficacy of our patient-centric model relies on data, creating a tension between improving MemoryCompanion's performance through more data and potential privacy breaches from data over-reliance. Striking an optimal balance is crucial. Exploring a phased data acquisition strategy with robust encryption and anonymization protocols is essential. Implementing a dynamic consent framework, allowing patients or guardians to control data sharing, can enhance protection.

\noindent\textbf{Authenticity of Representation} While current video-based interactions are innovative, there's room for improvement as demands for more immersive mediums grow. Future iterations may explore augmented reality (AR) or holographic representations for a lifelike and dynamic interaction. Collaboration with multimedia technology experts can guide the development of immersive patient experiences.

Resolving these limitations requires an approach centered around the patient, one that seamlessly incorporates ethical considerations and harnesses the potential of cutting-edge technological advancements.

\bibliography{jmlr-sample}

\newpage
\appendix

\section{MemoryCompanion Experiments Supplementary Video}\label{apd:first}
In this section, we present three experiments of MemoryCompanion across three distinct AD patients with different conditions. We provide concise introductions for each of the three AD patients as follows. The three provided instances exemplify the output of the MemoryCompanion pipeline for patients with varied demographic profiles. Comprehensive experiments are accessible within our pipeline repository on GitHub in Appendix \ref{apd:third}.

\begin{figure}[!ht]
\centering
\includegraphics[width=\linewidth]{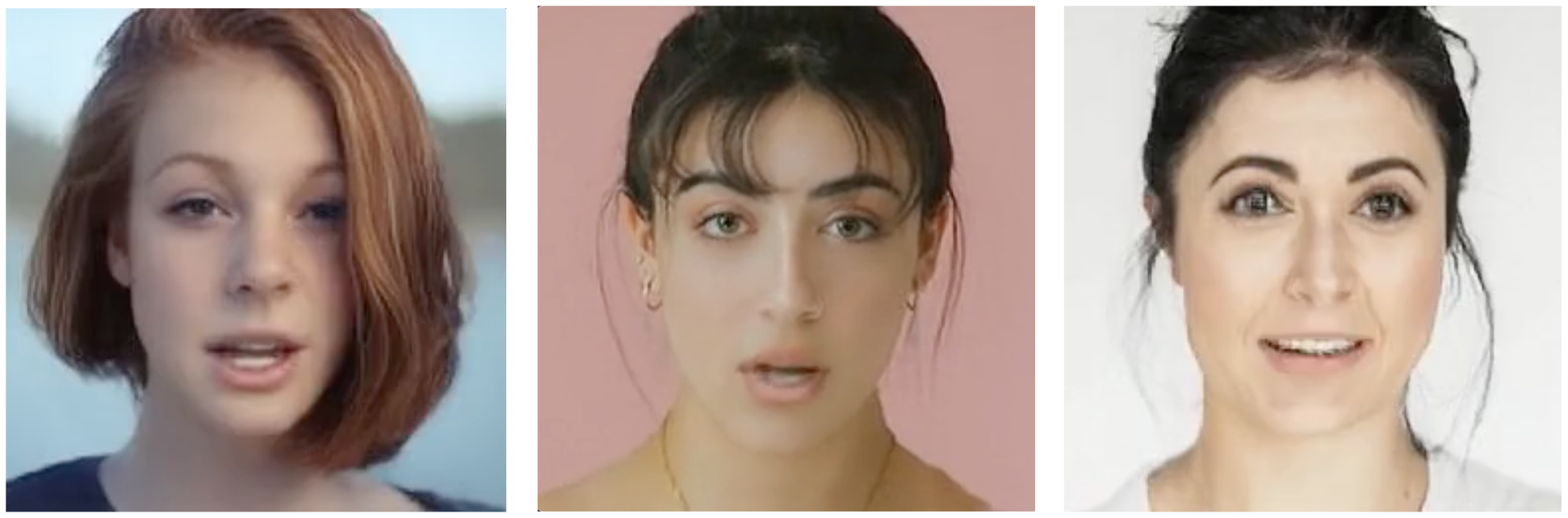}
\caption{MemoryCompanion Talking Interface. In the order from left to right, first is Margaret's MemoryCompanion (Amanda), second is David's MemoryCompanion (Eva), and third is Emily's MemoryCompanion (Julie). These talking face interfaces are sourced from ~\cite{unsplash}.}
\label{tf1}
\end{figure}

\textbf{Margaret Thompson}: A 92-year-old woman who dedicated her life to teaching. The MemoryCompanion tailored for Margaret incorporates the image of her favorite former student from her teaching career, Amanda. A visual representation of their interaction through the talking face interface is depicted in Figure \ref{tf1}. The full conversation in video between Margaret and MemoryCompanion (Amanda) can be found \href{https://drive.google.com/drive/folders/11gFR-UL1NyWgLkz96jqdlUhGWn_sclxl?usp=sharing}{here}. 

\textbf{David Johnson}: A 57-year-old who established himself as an entrepreneur. David's MemoryCompanion features his wife, Eva, as the conversational partner. A visual representation of their interaction through the talking face interface is depicted in Figure \ref{tf1}. The full conversation in video between David and MemoryCompanion (Eva) can be found \href{https://drive.google.com/drive/folders/1rUL8V9QMZ5dzYemZyGGaZpd8-z4eyAuF?usp=sharing}{here}.

\textbf{Emily Anderson}: A 24-year-old woman who recently completed her university education. Emily's MemoryCompanion features her closed friend, Julie. A visual representation of their interaction through the talking face interface is depicted in Figure \ref{tf1}. The full conversation in video between Emily and MemoryCompanion (Julie) can be found \href{https://drive.google.com/drive/folders/14KrGEd_FLDSBs1PZNhB0m3m0-RdyyX7A?usp=sharing}{here}.

\vspace{-0.3cm}
\section{Patient Synthetic Cases Dataset}\label{apd:second}
To foster transparency, reproducibility, and to offer a comprehensive dataset for further research, we are releasing our 10,000 synthetic patient profiles utilized in the MemoryCompanion research project. This \href{https://drive.google.com/drive/folders/1B1Qj6a5LY4wiOSGGqYv2yTZPGFe0ks5M?usp=sharing}{dataset} has been designed to emulate realistic patient profiles while ensuring no real patient data or personally identifiable information is included. It serves as a valuable resource for researchers and developers aiming to test, validate, or extend methods in related domains. We encourage researchers and practitioners to leverage this dataset for their experiments, ensuring that findings in this domain can be benchmarked and compared against a consistent data backdrop. This dataset is created by the following phases: 

\subsection{Collected Patient Profile Data}
We initially gathered genuine patient profile data from public resources, including Science, in addition to platforms such as the National Library of Medicine and the Department of Health~\cite{BCMJ, NLM, Science_case_study}. A representative real patient data case is showcased anonymously in Figure \ref{real_patient_data}. 

\subsection{Synthesized Patient Profile Data}
Based on the gathered data, we have formulated patient profile structures that encompass various aspects of each patient. This structure begins with demographic information such as age and ethnicity and further extends into medical history, familial background, and both professional and personal pursuits. Furthermore, it delves into the patient's present challenges, future aspirations, and emotional well-being. An illustrative synthesized case of the patient, Linda William, is presented below, in Figure \ref{synthesized_patient_case}.

\begin{figure}[!ht]
\begin{mdframed}[backgroundcolor=gray!20, linecolor=black]
\noindent\textbf{Name:} Mr. A. \\
\textbf{Age:} 68 years old \\
\textbf{Gender:} Male \\
\textbf{Medical Condition:} Cognitive decline \\
\textbf{Occupation:} Engineer \\
\textbf{Education:} 16 years \\
\textbf{Social History:} No recent history of alcohol or tobacco use \\
\textbf{Family History:} Mother had dementia, father had a heart attack \\
\textbf{Medical History:} High blood pressure, foot operation, stomach reflux \\
\textbf{Neurological Examination:} \\
- Mild difficulties in concentration and attention \\
- Language abnormalities: difficulty finding words, paraphasia \\
- Difficulty naming pictured objects (anomia) \\
- Repetition of complicated sentences \\
- Memory impairments \\
- Impaired arithmetic abilities \\
- Slight tremor in arms when outstretched \\
\noindent\textbf{Summary:} \\
Mr. A. showed difficulties in verbal communication and cognitive functions, primarily in language and memory. Despite these impairments, he continued working as an engineer and managed his daily activities with some support.
\end{mdframed}
\caption{Real Patient Profile of Mr. A, collected from Science~\cite{Science_case_study}}
\label{real_patient_data}
\end{figure}

\begin{figure}[h]
\begin{mdframed}[backgroundcolor=gray!20, linecolor=black]
\noindent\textbf{Name:} Linda Williams \\
\textbf{Gender:} Female \\
\textbf{Age:} 73 \\
\textbf{Ethnicity:} African American \\
\textbf{Religion:} Baptist \\
\textbf{Medical Condition:} Alzheimer’s \\
\textbf{First language:} English \\
\textbf{Family:} Son, brother \\
\textbf{Location:} Urban Atlanta \\
\noindent\textbf{Family Details:} \\
Son - Terrence, 49, a software engineer in San Francisco. \\
Brother - James, 75, a retired banker, lives in Florida. \\
\noindent\textbf{Summary:}\\
Linda was a former librarian. She lives in an apartment complex and has a close-knit group of friends from her church choir. She enjoys baking and often shares her treats with neighbors.
Recently, she's been forgetting song lyrics and has missed a few church services. Terrence is concerned as she once left the oven on after baking.\\
\noindent\textbf{What’s important to you?} \\
Linda values her independence and her choir practices. She loves sharing stories of her time as a librarian.\\
\noindent\textbf{What’s happening for you at the moment?} \\
Linda's friends have noticed she's less active in the choir and often repeats stories she has already told.\\
\noindent\textbf{What is the impact on you?} \\
She feels embarrassed when she forgets things, especially during choir practice.\\
\noindent\textbf{What would you like to happen in the future?} \\
She hopes to remain active in her community and maintain her current routines.\\
\noindent\textbf{How might we achieve this?} \\
Terrence is considering a move to a senior community where she can have more immediate assistance and supervision.\\
\noindent\textbf{What strengths and support networks do you have to help you?} \\
Linda's church community is supportive. Her son Terrence calls regularly and visits quarterly.\\
\end{mdframed}
\caption{Synthetic Patient Case Example}
\label{synthesized_patient_case}
\end{figure}

\section{MemoryCompanion Source Code on GitHub}\label{apd:third}
To champion transparency, reproducibility, and propel advancements in the field, we commit to releasing the entirety of the MemoryCompanion research source code on a GitHub Repository upon acceptance of the paper. This repository contains all essential components, scripts, and comprehensive documentation, simplifying replication and potential improvement of our work. We invite fellow researchers and enthusiasts to clone the repository, explore its structure and functionalities, and encourage contributions to this open-source initiative. Collaborative efforts hold the potential to drive significant progress in our field, and we're committed to fostering such collaboration through this release.

\end{document}